\documentclass{article}

\usepackage[preprint, nonatbib]{neurips_2020}

\usepackage[utf8]{inputenc} 
\usepackage{hyperref}       
\usepackage{url}            
\usepackage{amsmath}
\usepackage{multirow}
\usepackage{graphicx,wrapfig,lipsum}
\usepackage{amsfonts}
\usepackage{booktabs}
\usepackage{empheq}
\usepackage{rotating}
\usepackage{amsthm}
\usepackage{caption}
\usepackage{subcaption}
\usepackage{mathrsfs}  
\usepackage{amssymb}
\usepackage{tikz}
\usetikzlibrary{calc,positioning}

\theoremstyle{plain}

\theoremstyle{definition}

\newcommand{\set}[2]{\left\{#1\,\left\vert\,#2\vphantom{#1}\right\}\right.}
\newcommand{\reals}{\mathbb{R}}
\newcommand{\naturals}{\mathbb{N}}
\newcommand{\bigO}{\mathcal{O}}
\newcommand{\norm}[1]{\|#1\|}
\newcommand{\restr}[2]{{\left.\kern-\nulldelimiterspace #1 \right|_{#2}}}
\newcommand{\dee}{\,\mathrm{d}}
\newcommand{\logsig}{\mathrm{LogSig}}
\newcommand{\mfc}{\mathrm{MFC}}

\newcommand{\boldheading}[1]{

\textbf{#1}\quad}

\title{Generalised Interpretable Shapelets for Irregular Time Series}

\author{ 
	Patrick Kidger\thanks{Equal contribution.}
	\And
	James Morrill\footnotemark[1]
	\And
	Terry Lyons
	\AND \\[-12pt]
	Mathematical Institute, University of Oxford \\
	The Alan Turing Institute, British Library \\
	\texttt{\{kidger, morrill, tlyons\}@\hspace{0.8pt}maths.ox.ac.uk}
}

\begin{document}
	\maketitle
	\begin{abstract}
		The shapelet transform is a form of feature extraction for time series, in which a time series is described by its similarity to each of a collection of `shapelets'. However it has previously suffered from a number of limitations, such as being limited to regularly-spaced fully-observed time series, and having to choose between efficient training and interpretability. Here, we extend the method to continuous time, and in doing so handle the general case of irregularly-sampled partially-observed multivariate time series.  Furthermore, we show that a simple regularisation penalty may be used to train efficiently without sacrificing interpretability. The continuous-time formulation additionally allows for learning the length of each shapelet (previously a discrete object) in a differentiable manner. Finally, we demonstrate that the measure of similarity between time series may be generalised to a learnt pseudometric. We validate our method by demonstrating its performance and interpretability on several datasets; for example we discover (purely from data) that the digits 5 and 6 may be distinguished by the chirality of their bottom loop, and that a kind of spectral gap exists in spoken audio classification.
	\end{abstract}
	\section{Introduction}
	Shaplets are a form of feature extraction for time series \cite{ye2009firstshapelet, grabocka2014learningshapelet, hou2016efficient, bagnall2016bakeoff}. Given some fixed hyperparameter $K$, describing how many shapelets we are willing to consider, then each time series is represented by a vector of length $K$ describing how similar it is to each of the $k$ selected shapelets.
	
	We begin by recalling the classical definition of the shapelet transform \cite{hills2014classification}.
	\subsection{Classical shapelet transform}\label{section:classical-introduction}
	Given $N$ regularly sampled multivariate time series, with $D$ observed channels, where the $n$-th time series is of length $T_n$, then the $n$-th time series is a matrix 
	\begin{equation}\label{eq:f-n}
	f^n = (f^n_{t})_{t \in \{0, \ldots, T_n - 1\}} = (f^n_{t, d})_{t \in \{0, \ldots, T_n - 1\}, d \in \{1, \ldots, D\}},
	\end{equation}
	with each $f^n_{t, d} \in \reals$ and $n \in \{1, \ldots, N\}$.
	
	Fix some hyperparameter $K \in \naturals$, which will describe the number of shapelets. Fix some $S \in \{0, \ldots, \min_{i \in \{1, \ldots, N\}}T_n - 1\}$, which will describe the length of each shapelet. Then the $k$-th shapelet is a matrix
	\begin{equation*}
	w^{k} = (w^{k}_t)_{t \in \{0, \ldots, S - 1\}} = (w^{k}_{t, d})_{t \in \{0, \ldots, S - 1\}, d \in \{1, \ldots, D\}},
	\end{equation*}
	with each $w^{k}_{t, d} \in \reals$.
	
	Then the discrepancy between $f^n$ and $w^{k}$ is given by (sometimes without the square):
	\begin{equation}\label{eq:classical-shapelets}
	\sigma_S(f^n, w^{k}) = \min_{s \in \{0, \ldots, T_n - S\}} \sum_{t = 0}^{S - 1} \norm{f^n_{s + t} - w^{k}_t}_2^2,
	\end{equation}
	where $\norm{\,\cdot\,}_2$ describes the $L^2$ norm on $\reals^D$. A small discrepancy implies that $f^n$ and $w^{k}$ are similar to one another. This corresponds to sweeping $w^{k}$ over $f^n$, and finding the offset $s$ at which $w^{k}$ best matches $f^n$.
	
	In this article we will refer to the map
	\begin{equation}\label{eq:classical-shapelet-transform}
	f \mapsto (\sigma_S(f, w^{1}), \ldots, \sigma_S(f, w^{K}))
	\end{equation}
	as the \emph{classical shapelet transform}. The result is now a feature describing $f$, which may now be passed to some model to perform classification or regression.
	
	\subsection{Limitations}
	The classical shapelet method suffers from a number of limitations.
	\begin{enumerate}
	\item The technique only applies to regularly spaced time series.
	\item The choice of shapelet length $S$ is discrete and a hyperparameter. As such optimising it involves a relatively expensive hyperparameter search.
	\item Learning the shapelets $w^{k}$ by searching is expensive \cite{ye2009firstshapelet}, whilst optimising differentiably \cite{grabocka2014learningshapelet} typically sacrifices interpretability \cite{wang2019interp}.
	\end{enumerate}
	Besides this, the choice of $L^2$ norm is ad-hoc and a general formulation should allow for other notions of similarity. It is these limitations that we seek to address here.
	
	\subsection{Contributions}
	We extend the method to continuous time rather than discrete time. This allows for the treatment of irregularly-sampled partially-observed multivariate time series on the same footing as regular time series. Additionally, this continuous-time formulation means that the length of each shapelet (previously a discrete value) takes its values in a continuous range, and may now be trained differentiably.
	
	Next, we demonstrate how simple regularisation is enough to achieve shapelets that resemble characteristic features of the data. This gives interpretability with respect to a classification result, and also offers pattern discovery for determining previously unknown information about the data. For example we discover -- purely from data -- that the digits 5 and 6 may be distinguished by the chirality of their bottom loop, and that a kind of spectral gap exists in spoken audio classification.
	
	Finally, we generalise the discrepancy between a shapelet and a time series to be a learnt pseudometric. This is particularly useful for interpretability and pattern discovery, as doing so learns the importance of different channels.
	
	Our code is available at \texttt{https://github.com/patrick-kidger/generalised\_shapelets}.
	
	\section{Prior work}
	Shapelets may be selected as small intervals extracted from training samples \cite{ye2009firstshapelet}. However doing so is very expensive, requiring $\bigO(N^2 \cdot \max_n T_n^4)$ work. Much work on shapelets has sought speedup techniques \cite{mueen2011logical, grabocka2015scalable, grabocka2016speedshapelet}, for example via random algorithms \cite{rak2013fast, wistuba2015ultrafast}.
	
	However \cite{grabocka2014learningshapelet} observe that the discrepancy $\sigma_S$ of equation \eqref{eq:classical-shapelets} is differentiable with respect to $w^{k}$, so that shapelets may be differentiably optimised jointly with the subsequent model, as part of an end-to-end optimisation of the final loss function. (Although \cite{grabocka2014learningshapelet} include a `softmin' procedure which we believe to be unnecessary, as the minimum function is already almost everywhere differentiable.) This costs only $\bigO(N \cdot \max_n T_n^2)$ to train, and is the approach that we extend here.
	
	This method is attractive for its speed and its ease of trainability via modern deep learning frameworks \cite{tensorflow, pytorch, jax}. However, \cite{wang2019interp} observe that the predictive power of the distance between a shapelet and a time series need not correlate with a similarity between the two, so there is no pressure towards interpretability. \cite{wang2019interp} propose to solve this via adversarial regularisation; we will present a simpler alternative later. Without such procedures, then efficient training and interpretability become mutually exclusive.
	
	The method may additionally be generalised by considering alternative notions of similarity between a shapelet and a time series; for example \cite{grabocka2016dtwshapelet} replace the $L^2$ norm with dynamic time warping.
	
	The shapelet method is attractive for its normalisation of variable-length time series, and demonstratation of typically good performance \cite{bagnall2016bakeoff, bostrom2015shapelet}. Arguably its most important advantage is interpretability, as use of a particular feature corresponds to the importance of the similarity to the shapelet $w^{k}$. This may describe some shape that is characteristic of a particular class, and can discover previously unknown patterns in the data.
	
	\section{Method}
	\subsection{Continuous-time objects}
	We interpret a time series as a discretised sample from an underlying process, observed only through the time series. Similarly, a shapelet constructed as in Section \ref{section:classical-introduction} may be thought of as a discretisation of some underlying function. The first important step in our procedure is to construct continuous-time approximations to these underlying objects.
	
	\boldheading{Continuous-time path interpolants}
	Formally speaking, we assume that for $n \in \{1, \ldots, N\}$ indexing different time series, each of length $T_n$, we observe a collection of time series
	\begin{equation*}
	f^n = (f^n_{t_\tau})_{\tau \in \{1, \ldots, T_n\}},
	\end{equation*}
	where $t_\tau \in \reals$ is the observation time of $f^n_{t_\tau} \in (\reals \cup \{*\})^D$, where $*$ denotes the possiblity of a missing observation.
	
	Next, interpolate to get a function $\iota(f^n) \colon [0, T_n - 1] \to \reals^D$ such that $\iota(f^n)(t_\tau) = f^n_{t_\tau}$ for all $\tau \in \{0, \ldots, T_n - 1\}$ such that $f^n_{t_\tau}$ is observed. There are many possible choices for $\iota$, such as splines, kernel methods \cite{interpolation-prediction}, or Gaussian processes \cite{li2016scalable, futoma2017mgp}. In our experiments, we use piecewise linear interpolation.
	
	\boldheading{Continuous-time shapelets}
	The shapelets themselves we are free to control, and so for $k \in \{1, \ldots, K\}$ indexing different shapelets, we take each $w^{k, \rho} \colon [0, 1] \to \reals^D$ to be some learnt function depending on learnt parameters $\rho$. For example, this could be an interpolated sequence of learnt points, an expansion in some basis functions, or a neural network. In our experiments we use linear interpolation of a sequence of a learnt points.
	
	Then for some learnt length $S_k > 0$, define $w^{k, \rho, S_k} \colon [0, S_k] \to \reals^D$ by
	\begin{equation*}
	w^{k, \rho, S_k}(t) = w^{k, \rho}\left(\frac{t}{S_k}\right).
	\end{equation*}
	Taking the length $S_k$ to be continuous is a necessary prerequisite to training it differentiably. We will discuss the training procedure in a moment.
	
	\subsection{Generalised discrepancy}
	The core of the shapelet method is that the similarity or discrepancy between $f^n$ and $w^{k, \rho, S_k}$ is important. In general, we approach this by defining a \emph{discrepancy function} between the two, which will typically be learnt, and which we require only to be a pseudometric.
	
	We denote this discrepancy function by $\pi^A_{S}$. It depends upon a length $S$ and a learnt parameter $A$, consumes two paths $[0, S] \to \reals^D$, and returns a real number describing some notion of closeness between them. We are being deliberately vague about the regularity of the domain of $\pi^A_{S_k}$, as it is a function space whose regularity will depend on $\iota$.	
	
	Given some $\pi^A_{S}$, then the discrepancy between $f^n$ and $w^{k, \rho, S_k}$ is defined as
	\begin{equation}\label{eq:new-sigma}
	\sigma^A_{S_k}(f^n, w^{k, \rho, S_k}) = \min_{s \in [0, T_n - S_k]} \pi^A_{S_k}(\restr{\iota(f^n)}{[s, s + S_k]}(s + \cdot), w^{k, \rho, S_k}).
	\end{equation}
	
	The collection of discrepancies $(\sigma^A_{S_k}(f^n, w^{1, \rho, S_k}), \ldots, \sigma^A_{S_k}(f^n, w^{K, \rho, S_k}))$ is now a feature describing $f^n$, and is invariant to the length $T_n$. Use of the particular feature $\sigma^A_{S_k}(f^n, w^{k, \rho, S_k})$ corresponds to the importance of the similarity between $f^n$ and $w^{k, \rho, S_k}$. In this way, the choice of $\pi^A_{S_k}$ gives a great deal of flexibility.
	
	\boldheading{Existing shapelets fit into this framework}
	A simple example, in analogy to the classical shapelet method of equation \eqref{eq:classical-shapelets}, is to take
	\begin{equation*}
	\pi^A_{S_k}(f, w) = (\int_{0}^{S_k} \norm{f(t) - w(t)}_2^2 \dee t)^{\frac{1}{2}},
	\end{equation*}
	which in fact has no $A$ dependence. If $\iota$ is taken to be a piecewise constant `interpolation' then this will exactly correspond to (the square root of) the classical shapelet approach.
	
	\boldheading{Learnt $L^2$ discrepancies}
	The previous example may be generalised by taking our learnt parameter $A \in \reals^{D \times D}$, and then letting
	\begin{equation}\label{eq:learnt-discrepancy}
	\pi^A_{S}(f, w) = (\int_{0}^{S} \norm{A(f(t) - w(t))}_2^2 \dee t)^{\frac{1}{2}}.
	\end{equation}
	That is, allowing some learnt linear transformation before measuring the discrepancy. In this way, particularly informative dimensions may be emphasised. In our experiments we take $A$ to be diagonal. Allowing a general matrix was found during initial experiments to produce slightly worse performance.
	
	\boldheading{More complicated discrepancies}
	Moving on, we consider other more general choices of discrepancy, which may be motivated by the problem at hand. In particular we will discuss discrepancies based on the logsignature transform \cite{logsig-rnn}, and mel-frequency cepstrums (MFC) \cite{mfc}.
	
	Our exposition on these two discrepancies will be deliberately brief, as the finer details on exactly when and how to use them is domain-specific. The point is that our framework has the flexibility to consider general discrepancies motivated by other discplines, or which are known to extract information which is particular useful to the domain in question. An understanding of either logsignatures or mel-frequency cepstrums will not be necessary to follow the paper.
	
	\boldheading{Logsignature discrepancies}
	The logsignature transform is a transform on paths, known to characterise its input whilst extracting statistics which describe how the path controls differential equations \cite{logsig-rnn, levy-lyons, kidger2019deep, signatory, howisonutilisation}. Let $\mu$ denote the M{\"o}bius function, and let
	\begin{equation*}
	\beta_{D, R} = \sum_{r = 1}^R \frac{1}{r} \sum_{\rho \vert r} \mu\left(\frac{r}{\rho}\right) D^\rho,
	\end{equation*}
	which is Witt's formula \cite{witt}. Let
	\begin{equation*}
	\logsig^R \colon \set{f \colon [0, T] \to \reals^D}{T \in \reals, f\text{ is of bounded variation}} \to \reals^{\beta_{D, R}}
	\end{equation*}
	be the depth-$R$ logsignature transform. Let $A \in \reals^{\beta_{D, R} \times \beta_{D, R}}$ be full or diagonal as before, and let $\norm{\,\cdot\,}_p$ be the $L^p$ norm on $\reals^{\beta_{D, R}}$. Then we define the \emph{$p$-logsignature discrepancy} between two functions to be
	\begin{equation}\label{eq:logsignature-discrepancy}
	\pi^A_{S}(f, w) = \norm{A(\,\logsig^R(f) - \,\logsig^R(w))}_p.
	\end{equation}
	
	\boldheading{MFC discrepancies}
	The computation of an MFC is a function-to-function map derived from the short-time Fourier transform, with additional processing to focus on frequencies that are particularly relevant to human hearing \cite{mfc}. Composing this with the $L^2$ based discrepancy of equation \eqref{eq:learnt-discrepancy} produces
	\begin{equation}\label{eq:mfc-discrepancy}
	\pi^A_{S}(f, w) = (\int_{0}^{S} \norm{A(\mfc(f)(t) - \mfc(w)(t))}_2^2 \dee t)^{\frac{1}{2}}.
	\end{equation}
	
	\boldheading{The generalised shapelet transform}
	Whatever the choice of $\pi^A_S$, and in analogy to the classical shapelet transform \cite{hills2014classification}, we call the map
	\begin{equation}\label{eq:generalised-shapelet-transform}
	f \mapsto (\sigma^A_{S_1}(f, w^{1, \rho, S_1}), \ldots, \sigma^A_{S_K}(f, w^{K, \rho, S_K}))
	\end{equation}
	the \emph{generalised shapelet transform}.
	
	\subsection{Interpretable regularisation}
	As previously described, learning shapelets differentiably can sacrifice interpretability \cite{wang2019interp}, as the learnt shapelets need not resemble the training data. We propose a novel regularisation penalty to solve this: simply add on
	\begin{equation}
		\label{eq:interpretable_reg}
		\sum_{k = 1}^K \min_{n \in \{1, \ldots, N\}} \sigma^A_S(f^n, w^{k, \rho, s})
	\end{equation}
	as a regularisation term, so that minimising the discrepancy between $f^n$ and $w^{k, \rho, S}$ is also important. 
	Taking a minimum over $n$ asks that every shapelet should be similar to a single training sample, as in the original approach of finding shapelets by searching through the dataset instead of training differentiably.
	
	\subsection{Minimisation objective and training procedure}
	Overall, suppose we have some differentiable model $F^\theta$ parameterised by $\theta$, some loss function $\mathcal{L}$, and some observed time series $f^1, \ldots, f^N$ with targets $y_1, \ldots, y_N$.	
	
	Then letting $\gamma > 0$ control the amount of regularisation, we propose to seek a local minimum of
	\newcommand{\objective}{&\frac{1}{N}\sum_{n = 1}^N \mathcal{L}(y_n, F^\theta(\sigma^A_{S_1}(f^n, w^{1, \rho, S_1}), \ldots, \sigma^A_{S_K}(f^n, w^{K, \rho, S_K}))) + \gamma \sum_{k = 1}^K \min_{n \in \{1, \ldots, N\}} \sigma^A_{S_k}(f^n, w^{k, \rho, S_k})}
	\begin{align}
	\objective \label{eq:objective}
	\end{align}
	over model parameters $\theta$, discrepancy parameters $A$, shapelet parameters $\rho$, and shapelet lengths $S_k$, via standard stochastic gradient descent based techniques.

	\boldheading{Differentiability}
	Some thought is necessary to verify that this constructions is differentiable with respect to $S_k$. 
	There are two operations that may seem to pose a problem, namely the minimum over a range $\min_{s \in [0, T_n - S_k]}$, and the restriction operator $\iota(f^n) \mapsto \restr{\iota(f^n)}{[s, s + S_k]}$.

	Practically speaking, however, it is straightforward to resolve both of these issues. For the minimum over a range, this may reasonably be approximated by a minimum over some collection of points $s \in \{0, \varepsilon, 2 \varepsilon, \ldots, T_n - S_k - \varepsilon, T_n - S_k\}$, for some $\varepsilon > 0$ small and dividing $T_n - S_k$. This is now a standard piece of an autodifferentiation package. The error of this approximation may be controlled by the modulus of continuity of $s \mapsto \pi^A_{S_k}(\restr{\iota(f^n)}{[s, s + S_k]}(s + \cdot), w^{k, \rho, S_k})$, but in practice we found this to be unnecessary, and simply took $\varepsilon$ equal to the smallest gap between observations.
	
	Next, the continuous-time paths $\iota(f^n)$ and continuous-time shapelets $w^{k, \rho, S_k}$ must both be represented by some parameterisation of function space, and it is thus sufficient to restrict to considering differentiability with respect to this parameterisation.
	
	In our experiments we represent both $\iota(f^n)$ and $w^{k, \rho, S_k}$ as a continuous piecewise linear function stored as a collection of knots. In this context, the restriction operator is clearly differentiable, as a map from one collection of knots to a restricted collection of knots. Each knot is either kept (the identity function), thrown away (the zero function), or interpolated between to place a new knot at the boundary (a ratio of existing knots).

	\boldheading{Choice of $F^\theta$}\label{section:choice-of-f}
	Interpretability of the model will depend on an interpretable choice of $F^\theta$. In our experiments we thus used a linear model on the logarithm of every feature, so that a very negative coefficient corresponds to the importance of $f^n$ and $w^{k, \rho, S_k}$ being similar to each other.

\section{Experiments}\label{section:experiments}
We compare the generalised shapelet transform to the classical shapelet transform, in terms of both performance and interpretability, on a large range of time series classification problems. The shapelets of the classical shapelet transform are learnt differentiably.

In every case the model is a linear map, for interpretability as previously described, on either the generalised (equation \eqref{eq:generalised-shapelet-transform}) or classical (equation \eqref{eq:classical-shapelet-transform}) shapelet transforms. The learnt pseudometrics for the generalised shapelet transform scale each channel individually by taking $A$ to be diagonal.

Precise experimental details (optimiser, training scheme, ...) may be found in Appendix \ref{appendix:experimental}.

\subsection{The UEA Time Series Archive} \label{subsec:uea_classification}
This is a collection of 30 fully-observed regularly-sampled datasets with varying properties \cite{bagnall2018uea}, see Appendix \ref{appendix:experimental}. Evaluating on the full collection of datasets would take a prohibitively long time, and so we select 9 representing a range of difficulties.

We begin by performing hyperparameter optimisation 
for the classical shapelet transform, on each dataset. We then use the same hyperparameters for the generalised shapelet transform. For the generalised shapelet transform, the length hyperparameter is used to determine the initial length of the shapelet, but this may of course vary as it is learnt.

For the generalised shapelet transform, we consider two different discrepancy functions, specifically the learnt $L^2$ and $p$-logsignature discrepancies of equations \eqref{eq:learnt-discrepancy} and \eqref{eq:logsignature-discrepancy}. For the latter, we take $p=2$ and the depth $R=3$. We did not try to optimise $p$ and $R$, as we use the logsignature discrepancy simply to highlight the possibility of using more unusual discrepancies if desired.

\boldheading{Classification performance} The results are given in Table \ref{tab:uea_comparison_results}. We see that the generalised shapelet transform with $L^2$ discrepancy function achieves within one standard deviation of the top performing algorithm on 7 of the 9 datasets, whilst the classical approach does so for only 3.
\begin{table}[t]
    \centering
    \caption{Test accuracy (mean $\pm$ std, computed over three runs) on UEA. A `win' is the number of times each algorithm was within 1 standard deviation of the top performer for each dataset.}
    \label{tab:uea_comparison_results}
    \begin{tabular}{lccc}
\toprule
{} & \multicolumn{3}{c}{\textbf{Discrepancy}} \\ \cmidrule{2-4}
\textbf{Dataset} &          $L^2$ &   Logsignature &                Classical \\
\midrule
BasicMotions    &    90.8\% $\pm$ 1.4\% &  80.8\% $\pm$ 3.8\% &  \textbf{96.7\% $\pm$ 5.8\%} \\
ERing           &    \textbf{82.6\% $\pm$ 6.3\%} &  43.3\% $\pm$ 2.9\% &  67.2\% $\pm$ 11.8\% \\
Epilepsy        &    \textbf{88.4\% $\pm$ 3.0\%} &  \textbf{88.6\% $\pm$ 0.8\%} &  72.9\% $\pm$ 5.4\% \\
Handwriting     &    10.3\% $\pm$ 2.6\% &  \textbf{11.8\% $\pm$ 1.2\%} &  6.5\% $\pm$ 3.7\% \\
JapaneseVowels  &    \textbf{97.2\% $\pm$ 1.1\%} &  53.9\% $\pm$ 3.0\% &  91.5\% $\pm$ 4.1\% \\
Libras          &    \textbf{67.0\% $\pm$ 9.4\%} &  \textbf{67.8\% $\pm$ 5.5\%} &  62.2\% $\pm$ 2.4\% \\
LSST            &    \textbf{36.1\% $\pm$ 0.2\%} &  35.7\% $\pm$ 0.4\% &  33.5\% $\pm$ 0.5\% \\
PenDigits       &    \textbf{97.3\% $\pm$ 0.1\%} &  96.7\% $\pm$ 0.7\% &  \textbf{97.5\% $\pm$ 0.6\%} \\
RacketSports    &    \textbf{79.6\% $\pm$ 0.7\%} &  61.2\% $\pm$ 9.2\% &  \textbf{79.6\% $\pm$ 2.4\%} \\
\midrule
Wins &            7 &                 3 &    3 \\
\bottomrule
\end{tabular}

\end{table}

\begin{figure}[b]
\vspace{0.25em}
    \begin{subfigure}[b]{\linewidth}
        \centering
        \includegraphics[width=\linewidth]{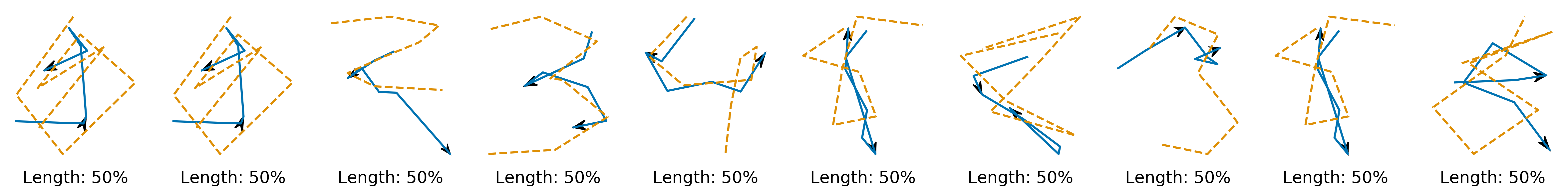}
        \caption{Classical shapelet transform.}
        \label{fig:old_shapelets}
    \end{subfigure}
    \begin{subfigure}[b]{\linewidth}
        \centering
        \includegraphics[width=\linewidth]{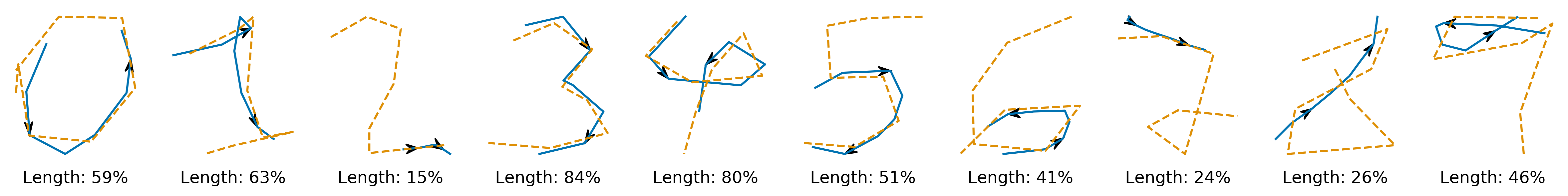}
        \caption{Generalised shapelet transform with $L^2$ discrepancy.}
        \label{fig:new_shapelets}
    \end{subfigure}
    \caption{The most significant shapelet for each class (blue, solid), overlaid with the most similar training example (orange, dashed). Similarity is measured with respect to the (learnt) discrepancy function.}
    \label{fig:pendigits}
\end{figure}

\boldheading{Interpretability on PenDigits}
We demonstrate interpretability by examining the PenDigits dataset. This is a dataset of handwritten digits 0--9, sampled at 8 points along their trajectory. We select the most informative shapelet for each of the ten classes (as in Section \ref{section:choice-of-f}), for both the classical shapelet transform and the generalised shapelet transform, with $L^2$ discrepancy. We then locate the training sample that it is most similar to, and plot an overlay of the two. See Figure \ref{fig:pendigits}.

We can clearly see multiple issues with the shapelets learnt with the classical approach. The most significant shapelet for the classes 0 and 1 is the same shapelet, and for classes 1, 5, 6, 7, 9, the most significant shapelet is not even closest to a member of that class. Visually, the shapelets for 3 and 4 seem to have identified distinguishing features of those classes, but the shapelets corresponding to the other classes appear to be little more than random noise.

In contrast, the results of the generalised shapelet approach are abundantly clear. Every class has a unique most significant shapelet, and every such shapelet is close to a member of the correct class. In the case of class 3, the shapelet has essentially reproduced the entire digit.


A point of interest is the difference between the shapelets for the digits 5 and 6, for the generalised shapelet transform. Whilst visually very similar, the difference between them is their direction. Whilst a 5 and a 6 may appear visually similar on the page (with a loop in the bottom half of the digit), they may clearly be distinguished by the direction in which they tend to be written. This is a nice example of discovering something about the data that was not necessarily already known!

Another such example is the shapelet corresponding to the class 7, for the generalised shapelet transform. This is perhaps surprising to see as a distinguising feature of a 7. However it turns out that no other digit uses a stroke in that direction, in that place! (Figuring this out was a fun moment for the authors, sketching figures in the air.) A similar case can be made for the 2 shapelet.

For further details see Appendix \ref{apx:uea}.

\subsection{Learning lengths, with irregularly sampled partially observed time series} \label{subsec:uea_missing_and_length}
We now investigate the strategy of learning lengths differentiably.

So as to keep things interesting, and to additionally provide benchmarks on irregularly sampled partially-observed datasets (to which the classical shapelet transform cannot be applied), for this test we drop either 10\%, 30\% or 50\% of the data for each of the JapaneseVowels, Libras and LSST datasets, selected for representing a range of difficulties. The data dropped is independently selected for every channel of each time series, and is the same for every model and repeat.

We use the generalised shapelet transform with learnt $L^2$ discrepancy, except we fix the lengths rather than learning them differentiably. We then perform a hyperparameter search to determine the best and worst lengths for this model on each dataset. We then train a model with differentiably learnt lengths initialised at the \emph{worst} lengths, and compare it to the \emph{best} performer from the hyperparameter search. See Table~\ref{tab:missing_and_length}. We see that the performance is comparable! This demonstrates that lengths learnt differentiably perform just as effectively as those selected by hyperparameter search, but without the relatively more expensive search.

\begin{table}[t]
    \centering
	\caption{Test accuracy (mean $\pm$ std, computed over three runs) on three UEA datasets with missing data. A `win' is defined as the number of times each algorithm was within 1 standard deviation of the top performer for each dataset.}\label{tab:missing_and_length}
    \begin{tabular}{lccc}
\toprule
& {} & \multicolumn{2}{c}{\textbf{Lengths selected by}} \\  \cmidrule{3-4}
\textbf{Dataset} & \parbox{20mm}{\centering\textbf{Dropped data}} & \parbox{20mm}{\centering Differentiable optimisation} & \parbox{20mm}{\centering Hyperparameter searching} \\
\midrule
\multirow{3}{*}{JapaneseVowels} & 10\% & \textbf{93.2\% $\pm$ 2.1\%} & \textbf{93.1\% $\pm$ 0.9\%} \\
                                & 30\% & \textbf{91.2\% $\pm$ 4.1\%} & \textbf{91.4\% $\pm$ 2.8\%} \\
                                & 50\% & \textbf{93.5\% $\pm$ 1.1\%} & 92.0\% $\pm$ 1.1\% \\
\multirow{3}{*}{Libras}         & 10\% & 57.4\% $\pm$ 4.2\% & \textbf{59.3\% $\pm$ 1.6\%} \\
                                & 30\% & \textbf{81.2\% $\pm$ 7.6\%} & 63.9\% $\pm$ 8.2\% \\
                                & 50\% & \textbf{62.5\% $\pm$ 14.8\%} & \textbf{65.3\% $\pm$ 8.6\%} \\
\multirow{3}{*}{LSST}           & 10\% & 40.2\% $\pm$ 3.5\% & \textbf{44.0\% $\pm$ 1.0\%} \\
                                & 30\% & \textbf{38.1\% $\pm$ 0.3\%} & \textbf{40.2\% $\pm$ 5.6\%} \\
                                & 50\% & 41.5\% $\pm$ 2.7\% & \textbf{44.4\% $\pm$ 0.5\%} \\
\midrule
Wins                            &    & 6 & 7 \\
\bottomrule
\end{tabular}

\end{table}

For further details see Appendix \ref{apx:learninglengths}.

\subsection{Speech Commands}
Finally we consider the Speech Commands dataset~\cite{warden2018speech}. This is comprised of one-second audio files, corresponding to words such as `yes', `no', `left', `right', and so on. We consider 10 classes so as to create a balanced classification problem.

For the generalised shapelet transform, we use the MFC discrepancy described in equation~\eqref{eq:mfc-discrepancy}.

\boldheading{Classification performance} For this more difficult dataset, the generalised shapelet transform substantially outperformed the classical shapelet transform. (To keep things fair, the classical shapelet transform is used in MFC-space; the performance gap is not due to this.) The classical shapelet transform produces a test accuracy of 44.8\% $\pm$ 8.6\%, whilst the generalised shapelet transform produces a test accuracy of 91.9\% $\pm$ 2.4\% (mean $\pm$ std, averaged over three runs).

\boldheading{Interpretability}

\begin{figure}
			\centering
			\includegraphics[width=\linewidth]{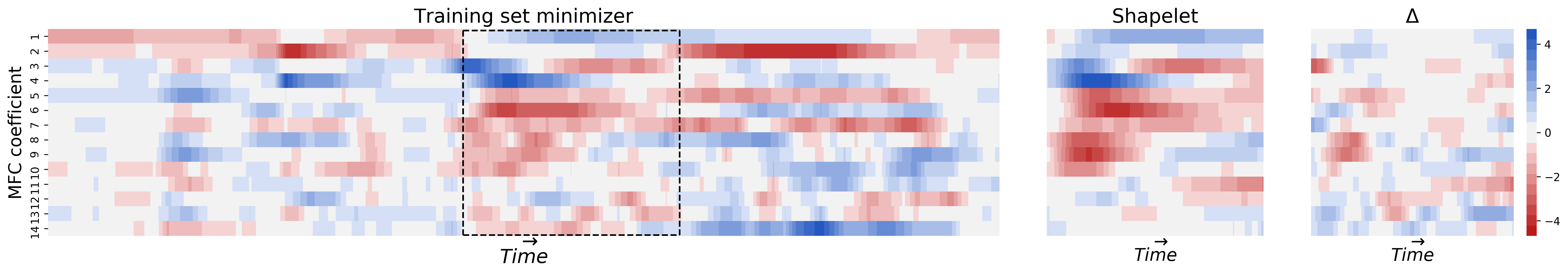}
			\caption{Generalised shapelet transform with learnt $L^2$ discrepancy. First 15 MFC coefficients for the training set minimizer (left), shapelet (middle), and the difference between them (right). The dashed box in the minimizer plot indicates the position in the series that the shapelet corresponds to.}\label{fig:speech_commands}
\end{figure}

We examine interpretability in three different ways. First we consider MFC-space, see Figure \ref{fig:speech_commands}. We see that the shapelets have learnt to resemble small segments of the training data, so that classification may be determined by the presence of different frequencies.

Furthermore, these shapelets may be listened to as audio! The audio files may be found at \texttt{https://github.com/patrick-kidger/generalised\_shapelets/tree/master/audio}. The audio to MFC map is naturally lossy, so the shapelets are far from perfect, but the difference between them is nonetheless clear. The shapelet most strongly associated with `left' captures the `eft' sound, whilst the one for `stop' actually sounds like the word itself. Much like the shapelet associated with class 7 in the PenDigits example, the sounds extracted need not resemble the word in isolation. Instead, they capture features that distinguishes that class from the others present.

\begin{wrapfigure}[10]{l}{5cm}\centering
\vspace{-2em}
\includegraphics[width=\linewidth]{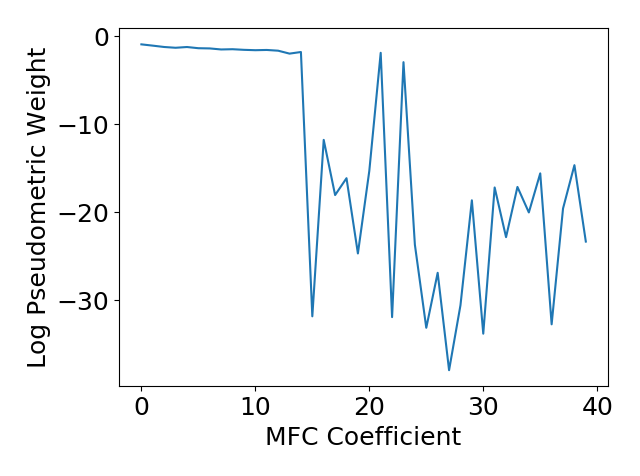}
\vspace{-2em}
\caption{Pseudometric channel weighting identifies a spectral gap.}\label{fig:spectral}
\end{wrapfigure}

Finally, we examine the coefficients of the learnt $L^2$ pseudometric, recalling that the matrix $A$ of equation \eqref{eq:mfc-discrepancy} is diagonal and thus weights the importance of each channel. See Figure \ref{fig:spectral}. The coefficients of the pseudometric have learnt to be relatively large for the first 15 channels, and dramatically smaller for the later 25 channels. The pseudometric has learnt -- purely from data -- a quantitative description of the fact that lower frequencies are more important to distinguish words \cite{monson2019maximum}. In short, it has discovered a kind of spectral gap! 

See Appendix \ref{apx:further_speech_commands_interpretability} for further plots.


	\section{Conclusion}
	In this work we have generalised the classical shapelet method in several ways. We have generalised it from discrete time to continuous time, and in doing so extended the method to the general case of irregularly-sampled partially-observed multivariate time series. Furthermore this allows for the length of each shapelet to be treated as a parameter rather than a hyperparameter, and optimised differentiably. We have introduced generalised discrepancies to allow for domain adaptation. Finally we have introduced a simple regularisation penalty that produces interpretable results capable of giving new insight into the data.
	
	\section*{Broader Impact}
	Interpretability is important in the application of many machine learning systems, often over and above raw performance, so that the reason for choices made on the basis of that system can be justified, seen to be made fairly, and without undue bias. Furthermore, methods which give new insight into the data are valuable for their ability to help the subequent development of theory. The generalised shapelet transform, with interpretable regularisation, is capable of supporting both of these objectives, and so it is our hope that a substantial part of the broader impact of this work will be its contributions towards these strategic goals.

	\begin{ack}
	PK was supported by the EPSRC grant EP/L015811/1. JM was supported by the EPSRC grant EP/L015803/1 in collaboration with Iterex Therapuetics. PK, JM, TL were supported by the Alan Turing Institute under the EPSRC grant EP/N510129/1.
	\end{ack}
	
	\small
	\bibliographystyle{ieeetr}
	\bibliography{references} 
	
	\normalsize
	\newpage
	\appendix
	
	\section{Experimental details}\label{appendix:experimental}
	\subsection{General notes}
	Many details of the experiments are already specified in Section \ref{section:experiments}, and we do not repeat those details here.
	
	\boldheading{Code} Code to reproduce every experiment can found at \texttt{https://github.com/patrick-kidger/generalised\_shapelets}.
	
	\boldheading{Choice of $\iota$} The interpolation scheme $\iota$ is taken to be piecewise linear interpolation. In particular efficient algorithms for computing the logsignature transform only exist for piecewise linear paths \cite{signatory}.
	
	\boldheading{Regularisation parameter} The parameter $\gamma$ for the interpretable regularisation is taken to be $10^{-4}$. This was selected by starting at $10^{-3}$ and reducing the value until test accuracy no longer improved, so as to ensure that it did not compromise performance.
	
	\boldheading{Optimisation} 
The loss was cross entropy, the optimiser was Adam \cite{kingma2015} with learning rate 0.05 and batch size 1024. If validation loss stagnated for 20 epochs then the learning rate was reduced by a factor of 10 and training resumed, down to a minimum learning rate of 0.001. We note that these relatively large learning rates are (as is standard practice) proportional to the large batch size. If validation loss and accuracy failed to decrease over 60 epochs then training was halted. Once training was completed then the model parameters were rolled back to those which produced the highest validation accuracy.
	
	\boldheading{Computer infrastructure} Experiments were run on the CPU of a variety of different machines, all using Ubuntu 18.04 LTS, and running PyTorch 1.3.1.
	
	\subsection{UEA} \label{apx:uea}
	The datasets can be downloaded from \texttt{https://timeseriesclassification.com}.

	The maximum number of epochs allowed for training was 250.	

	All UEA datasets were used unnormalised. Those samples which were shorter than the maximum length of the sequence were padded to the maximum length by repeating their final entry.
	
	Hyperparameters were found by performing a grid search over 2, 3, 5 shapelets \emph{per class}, with a maximum total number of shapelets of 30, and shapelets being set to (classical shapelet transform) / initialised at (generalised shapelet transform) 0.15, 0.3, 0.5, 1.0 times the maximum length of the time series.
	
	The dataset comes with default train/test splits, which we respect here. The training data is split 80\%/20\% into train and validation sets, stratified by class label.
	
	The details of each dataset are as below. We note that the train/test splits are sometimes of unusual proportion; we do not know the reason for this odd choice.

	\begin{table}[ht]
		\caption{UEA dataset details and hyperparameter choices}
		\label{tab:uea_hyperparams_old}
		\centering
		\begin{tabular}{lccccccc}
			\toprule
			Dataset & Train size & Test size & Dimensions & Length & Classes & \parbox{10mm}{Shapelets} & \parbox{10mm}{Shapelet length fraction} \\ \midrule
			BasicMotions    & 40 & 40 & 6 & 100 & 4 & 12 & 0.5 \\
			ERing           & 30 & 30 & 4 & 65 & 6 & 12 & 0.5 \\
			Epilepsy        & 137 & 138 & 3 & 206 & 4 & 20 & 0.5 \\
			Handwriting     & 150 & 850 & 3 & 152 & 26 & 30 & 0.5 \\
			JapaneseVowels  & 270 & 370 & 12 & 29 & 9 & 18 & 0.5 \\
			Libras          & 180 & 180 & 2 & 45 & 15 & 30 & 1.0 \\
			LSST            & 2459 & 2466 & 6 & 36 & 14 & 28 & 1.0 \\
			PenDigits       & 7494 & 3498 & 2 & 8 & 10 & 30 & 0.5 \\
			RacketSports    & 151 & 152 & 6 & 30 & 4 & 12 & 0.5 \\
			\bottomrule
		\end{tabular}
	\end{table}
	
	\subsection{Learning lengths}\label{apx:learninglengths}

	Table \ref{tab:uea_hyperparams_l2} shows the hyperparameters used in Section \ref{subsec:uea_missing_and_length}. These were chosen to optimize the validation score for the generalised shapelet transform without learnt lengths, rather than the classical shapelet transform, and as such are different to those noted above. The hyperparameters were optimised for the 30\% drop rate, and the same hyperparameters simply used for the 10\% and 50\% drop rate cases.

	\begin{table}[ht]
		\caption{Hyperparameter choices for study on learnt lengths}
		\label{tab:uea_hyperparams_l2}
		\centering
		\begin{tabular}{lccc}
			\toprule
			Dataset & Shapelets & Best length fraction & Worst length fraction \\
			\midrule
			JapaneseVowels  & 27 & 0.15 & 0.5 \\
			Libras          & 30 & 1.0 & 0.15 \\
			LSST            & 28 & 0.3 & 1.0 \\
			\bottomrule
		\end{tabular}
	\end{table}
	
	\subsection{Full hyperparameter search results}
	For completeness we also give the full results from both hyperparameter searches in the preceding two sections, in Tables \ref{tab:uea_hyperparams_old_results} and \ref{tab:uea_hyperparams_l2_results}.

	\begin{sidewaystable}
		\centering
		\begin{tabular}{lcccccccccccc}
\toprule
& \multicolumn{12}{c}{\textbf{Hyperparameter options.} Top row: shapelets per class. Bottom row: Shapelet length fraction.}\\
{}                        & \multicolumn{4}{c}{{2}}    & \multicolumn{4}{c}{{3}}    & \multicolumn{4}{c}{{5}}  \\ \cmidrule(lr){2-5} \cmidrule(lr){6-9} \cmidrule(lr){10-13}
\textbf{Dataset}          & 0.15   & 0.3    & 0.5    & 1.0    & 0.15   & 0.3    & 0.5    & 1.0    & 0.15   & 0.3    & 0.5    & 1.0 \\
\midrule
BasicMotions              &                100.0\% &     100.0\% &     100.0\% &     100.0\% &      100.0\% &     \textbf{100.0\%} &     100.0\% &      75.0\% &      100.0\% &     100.0\% &     100.0\% &     100.0\% \\
ERing                     &                 83.3\% &     100.0\% &     \textbf{100.0\%} &     100.0\% &      100.0\% &      83.3\% &     100.0\% &     100.0\% &      100.0\% &      83.3\% &      83.3\% &     100.0\% \\
Epilepsy                  &                 82.1\% &      75.0\% &      82.1\% &      82.1\% &       82.1\% &      85.7\% &      82.1\% &      71.4\% &       82.1\% &      82.1\% &      \textbf{89.3\%} &      75.4\% \\
Handwriting               &                 13.3\% &      16.7\% &      23.3\% &      23.3\% &       16.7\% &      23.3\% &      \textbf{26.7\%} &      23.3\% &       16.7\% &      20.0\% &      16.7\% &      23.3\% \\
JapaneseVowels            &                 90.7\% &      90.7\% &      \textbf{96.3\%} &      92.6\% &       92.6\% &      92.6\% &      94.4\% &      92.6\% &       90.7\% &      92.6\% &      96.3\% &      90.7\% \\
Libras                    &                 83.3\% &      80.6\% &      88.9\% &      83.3\% &       77.8\% &      88.9\% &      86.1\% &      77.8\% &       77.8\% &      86.1\% &      80.6\% &      \textbf{91.7\%} \\
LSST                      &                 34.8\% &      33.7\% &      33.3\% &      \textbf{35.2\%} &       33.7\% &      33.9\% &      35.0\% &      34.1\% &       33.7\% &      33.7\% &      35.0\% &      34.6\% \\
PenDigits                 &                 97.6\% &      98.0\% &      98.7\% &      96.7\% &       98.0\% &      97.9\% &      98.4\% &      96.5\% &       97.6\% &      98.5\% &      \textbf{98.9\%} &      96.3\% \\
RacketSports              &                 61.3\% &      74.2\% &      83.9\% &      80.6\% &       58.1\% &      67.7\% &      \textbf{87.1\%} &      77.4\% &       71.0\% &      77.4\% &      64.5\% &      80.6\% \\
\bottomrule
\end{tabular}

		\caption{Accuracy on the validation set for the hyperparameter runs performed to determine the hyperparameters used in Section \ref{subsec:uea_classification}. The upper row represents the number of shapelets per class, with the lower row being the shapelet length fraction. The best run is given in bold. When multiple options achieved the highest score, the hyperparameters were chosen randomly from that top performing set. Only one run was performed for each hyperparameter option.}\label{tab:uea_hyperparams_old_results}

		\bigskip\bigskip  

		\centering
		\begin{tabular}{lcccccccccccc}
\toprule
& \multicolumn{12}{c}{\textbf{Hyperparameter options.} Top row: shapelets per class. Bottom row: Shapelet length fraction.}\\
{} & \multicolumn{4}{c}{{2}}    & \multicolumn{4}{c}{{3}}    & \multicolumn{4}{c}{{5}}  \\
\cmidrule(lr){2-5} \cmidrule(lr){6-9} \cmidrule(lr){10-13}
\textbf{Dataset}          & 0.15   & 0.3    & 0.5    & 1.0    & 0.15   & 0.3    & 0.5    & 1.0    & 0.15   & 0.3    & 0.5    & 1.0 \\
\midrule
JapaneseVowels            & 94.4\% & 94.4\% & 94.4\% & 92.6\% & \textbf{96.3\%}$^*$ & \textbf{94.4\%}$_*$ & 94.4\% & 94.4\% & 96.3\% & 94.4\% & 92.6\% & 92.6\% \\
Libras                    & \textbf{63.9\%}$_*$ & 77.8\% & 77.8\% & \textbf{88.9\%}$^*$ & 75.0\% & 80.6\% & 83.3\% & 83.3\% & 73.2\% & 69.4\% & 83.3\% & 86.1\% \\
LSST                      & 45.7\% & \textbf{46.5\%}$^*$ & 42.5\% & \textbf{37.0\%}$_*$ & 41.9\% & 43.3\% & 42.6\% & 39.0\% & 45.5\% & 43.5\% & 41.4\% & 39.4\% \\
\bottomrule
\end{tabular}

		\caption{Accuracy on the validation set for the hyperparameter runs performed to determine the hyperparameters used in Section \ref{subsec:uea_missing_and_length}. The upper row represents the number of shapelets per class, with the lower row being the shapelet length fraction. The best and worst hyperparameters -- recall that worst is also used here -- are denoted in bold. The best case additionally has a superscript $^*$ and the worst case additionally has a superscript $_*$. When multiple options achieved the highest score, the hyperparameters were chosen randomly from that top performing set.}\label{tab:uea_hyperparams_l2_results}
	\end{sidewaystable}

	\subsection{Speech Commands}\label{apx:further_speech_commands_interpretability}
	The dataset can be downloaded from\\ \texttt{http://download.tensorflow.org/data/speech\_commands\_v0.02.tar.gz}.
	
	We began by selecting every sample from the `yes', `no', `up', `down', `left', `right', `on', `off', `stop' and `go' categories, and discarding the samples which were not of the maximum length (16000; nearly every sample is of this maximum length). This gives a total of 34975 samples.
	
	The samples were preprocessed by computing the MFC with a Hann window of length 400, hop length 200, 400 frequency bins, 128 mels, and 40 MFC coefficients. Every sample is then of length 81 with 40 channels. Every channel was then normalised to have mean zero and variance one.
	
	No hyperparameter searching was performed, due to the inordinately high cost of doing so - shapelets are an expensive algorithm that is primarily a `small data' technique, and this represented the upper limit of problem size that we could consider! That said, this is in large part an implementation issue. An efficient GPU implementation should be possible. The problem is that current machine learning frameworks \cite{tensorflow, pytorch, jax} typically parallelise by vectorising every operation, however this problem (which is embarrasingly parallel) is instead best handled via na{\"i}ve parallelism at the top level. This is because of the need for different behaviour for different batch elements; they will in general have minimisers at different sections of the time series, meaning that a vectorised approach needs to keep track of the union of these points for every batch element. We highlight that not needing to perform hyperparameter optimisation on the length is an advantage of our generalised shapelet transform, thus reducing this kind of computational burden.
	
	The maximum number of epochs allowed for training was 1000. The number of shapelets used per class was 4, for a total of 40 shapelets. The length of each shapelet (set to for the classical shapelet transform; initialised at for the generalised shapelet transform) was taken to be 0.3 of the full length of the dataset.
	
	The data is combined into a single dataset and a 70\%/15\%/15\% training/validation/test split taken, stratified by class label.

\begin{figure}[t]
			\centering
			\includegraphics[width=\linewidth]{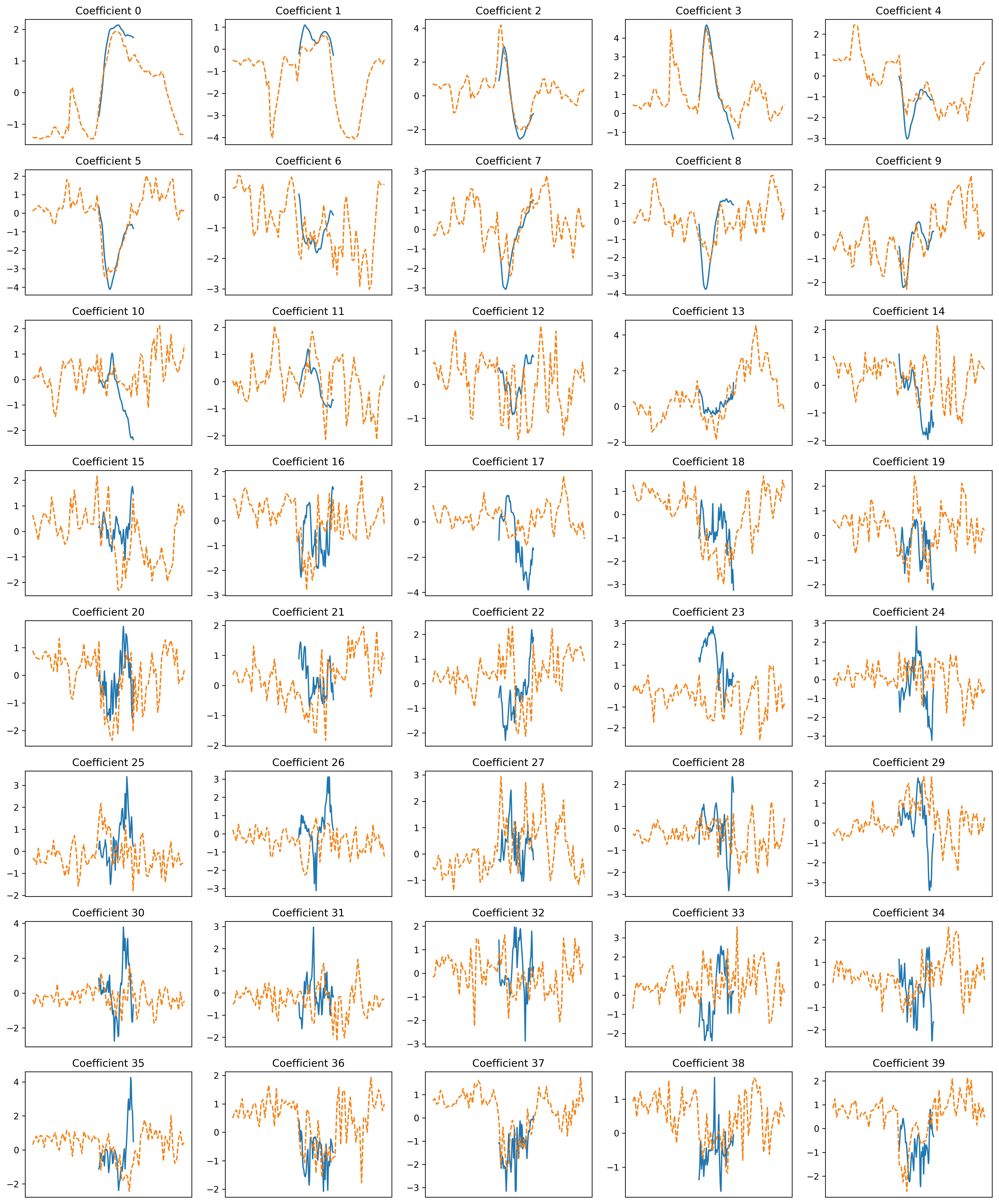}
		\caption{Generalised shapelet transform with $L^2$ discrepancy. All MFC coefficients for the shapelet (blue) and the training set minimizer (orange, dashed).}
		\label{fig:speech_commands_axisplot}
	\end{figure}

	In Figure \ref{fig:speech_commands_axisplot} we show all 40 MFC coefficients for the shapelet (blue) and the training set minimizer (orange, dashed) for the generalised shapelet transform with $L^2$ discrepancy, for a particular (arbitrarily selected) run.

\end{document}